\documentclass[sn-mathphys]{sn-jnl}
\usepackage{fix-cm}
\jyear{2021}
\usepackage{bm}
\usepackage{url}
\usepackage{xcolor}
\usepackage{wrapfig}
\usepackage{comment}
\usepackage{appendix}
\usepackage{booktabs}
\usepackage{multirow}
\usepackage{enumitem}
\usepackage{hyperref}
\usepackage{graphicx}
\usepackage {makecell}
\usepackage{algorithm}
\usepackage{subcaption}
\usepackage{algpseudocode}
\usepackage{threeparttable}
\usepackage{amsmath,amssymb,amsfonts}

\begin{document}

\title[RL-MolGAN: RL-based Transformer GAN for Molecular Generation]{A Reinforcement Learning-Driven Transformer GAN for Molecular Generation}

\author*[1]{\fnm{Chen} \sur{Li}}\email{li.chen@ist.osaka-u.ac.jp}
\author[2]{\fnm{Huidong} \sur{Tang}}\email{tanghd24@163.com}
\author[3]{\fnm{Ye} \sur{Zhu}}\email{ye.zhu@ieee.org}
\author[4]{\fnm{Yoshihiro} \sur{Yamanishi}}\email{yamanishi@i.nagoya-u.ac.jp}

\affil[1]{\orgdiv{D3 Center}, \orgname{Osaka University}, \orgaddress{\city{Osaka} \postcode{5670047}, \country{Japan}}}
\affil[2]{\orgdiv{Graduate School of Advanced Science and Engineering}, \orgname{Hiroshima University}, \orgaddress{\city{Higashi-Hiroshima} \postcode{7398521}, \country{Japan}}}
\affil[3]{\orgdiv{Centre for Cyber Resilience and Trust}, \orgname{Deakin University}, \orgaddress{\city{Burwood} \postcode{3125},  \country{Australia}}}
\affil[4]{\orgdiv{Graduate School of Informatics}, \orgname{Nagoya University}, \orgaddress{\city{Nagoya} \postcode{4648602},  \country{Japan}}}
 
\abstract{Generating molecules with desired chemical properties presents a critical challenge in fields such as chemical synthesis and drug discovery. Recent advancements in artificial intelligence (AI) and deep learning have significantly contributed to data-driven molecular generation. However, challenges persist due to the inherent sensitivity of simplified molecular input line entry system (SMILES) representations and the difficulties in applying generative adversarial networks (GANs) to discrete data. This study introduces RL-MolGAN, a novel Transformer-based discrete GAN framework designed to address these challenges. Unlike traditional Transformer architectures, RL-MolGAN utilizes a first-decoder-then-encoder structure, facilitating the generation of drug-like molecules from both $de~novo$ and scaffold-based designs. In addition, RL-MolGAN integrates reinforcement learning (RL) and Monte Carlo tree search (MCTS) techniques to enhance the stability of GAN training and optimize the chemical properties of the generated molecules. To further improve the model's performance, RL-MolWGAN, an extension of RL-MolGAN, incorporates Wasserstein distance and mini-batch discrimination, which together enhance the stability of the GAN. Experimental results on two widely used molecular datasets, QM9 and ZINC, validate the effectiveness of our models in generating high-quality molecular structures with diverse and desirable chemical properties.}

\maketitle

\section{Introduction}
\label{sec:introduction}

In silico generation of molecules with desired chemical properties has emerged as a critical area in chemical synthesis \cite{struble2020current} and drug discovery \cite{meyers2021novo}. Traditional computational methods, such as molecular docking, while insightful, often lack the efficiency and accuracy required for high-throughput applications, especially when optimizing for specific properties like drug-likeness or solubility \cite{kumari2023synthesis}. Recent advances in artificial intelligence (AI) and deep learning have enabled data-driven molecular generation, with models such as variational autoencoders (VAEs) \cite{jin2018junction,li2024gxvaes}, generative adversarial networks (GANs) \cite{goodfellow2014generative,li2022transformer}, and, more recently, stable diffusion models \cite{xu2023geometric} being explored for generating molecular structures. These structures are often represented in formats like simplified molecular input line entry system (SMILES) strings \cite{weininger1988smiles}, self-referencing embedded strings (SELFIES) \cite{krenn2020self}, or as molecular graphs \cite{liu2019n}.

SMILES, one of the most widely used notations, encodes molecules as text-like strings that represent atom connectivity in a linear format. However, generating accurate SMILES strings can be challenging due to their sensitivity to small perturbations, which can result in syntactically correct but chemically invalid molecules \cite{li2024tengan}. SELFIES overcomes some of the limitations of SMILES by introducing a syntax that guarantees the chemical validity of generated molecules, making it particularly useful for tasks where chemical correctness is essential \cite{krenn2022selfies}. However, the complex semantics of SELFIES mean that even minor syntactic changes can lead to molecular structures with significant differences in chemical composition, functional groups, and other key properties \cite{vogt2023exploring}. In contrast, molecular graphs explicitly represent the connectivity and relationships between atoms and bonds, offering a more detailed depiction of molecular structure \cite{liu2019n}. Although highly effective for tasks requiring structural fidelity, molecular graphs are often complex and computationally intensive, limiting their scalability in high-throughput applications \cite{sun2020graph}.

Due to their simplicity and suitability for processing by natural language processing (NLP) deep learning models in a text-like manner, SMILES representations have become the most widely used format for molecular generation tasks \cite{mswahili2024transformer}. In NLP, the Transformer model has proven highly effective at capturing global dependencies and long-range relationships within sequential data \cite{vaswani2017attention}. Its ability to process all parts of an input sequence simultaneously makes it especially effective at addressing the sensitivity of SMILES representations to small perturbations \cite{arus2019randomized}. For instance, the self-attention mechanism in the Transformer enables each atom in a SMILES string to interact with every other atom, creating a global receptive field that is crucial for generating chemically valid molecular structures. However, using the Transformer to generate molecules based on a fixed scaffold typically poses two main challenges \cite{li2023spotgan}. First, the Transformer's positional encoding fails to preserve the positions of attachment points on the scaffold, which are critical for determining where functional groups should be attached. Second, a functional group may consist of a variable number of atoms, while the Transformer is not designed to handle sequences of varying lengths. Addressing these challenges is crucial to ensuring that the generated molecules maintain structural consistency with the fixed scaffold.

On the other hand, GAN models, originally developed for continuous data such as image generation \cite{goodfellow2014generative}, encounter unique challenges when applied to discrete molecular representations like SMILES strings \cite{yu2017seqgan}. Traditional GANs are designed for smooth output spaces, where continuous gradients enable effective adversarial training \cite{gulrajani2017improved}. In contrast, the discrete nature of SMILES strings disrupts this process: because SMILES data lacks continuity, the gradient-based optimization essential for GAN training becomes less effective, often leading to unstable training and inconsistent outputs \cite{guimaraes2017objective}. Additionally, for discrete data, adversarial feedback in GANs is only available after the entire SMILES string is generated, making it difficult to refine the output iteratively during the generation process \cite{li2022transformer}.

To tackle these challenges, this study introduces RL-MolGAN, a novel Transformer-based discrete GAN framework that utilizes a first-decoder-then-encoder structure, deviating from the traditional first-encoder-then-decoder approach. This design choice enhances the framework's capability to generate molecular structures more effectively. Specifically, RL-MolGAN employs a Transformer decoder variant as the generator to produce SMILES strings, while the discriminator, based on a Transformer encoder variant, maintains a global receptive field to steer the generator towards producing drug-like molecules. Additionally, RL-MolGAN integrates reinforcement learning (RL) \cite{kimura1995reinforcement} with Monte Carlo tree search (MCTS) \cite{li2022transformer} to stabilize GAN training and optimize the chemical properties of the generated SMILES strings. This approach leverages the power of RL to handle discrete data spaces, ensuring better training stability and overcoming the inherent challenges of optimizing discrete molecular structures. Moreover, to further enhance the stability and performance of the framework, an extended version, RL-MolWGAN, is proposed. RL-MolWGAN incorporates the Wasserstein distance \cite{arjovsky2017wasserstein} and mini-batch discrimination \cite{salimans2016improved} to improve the training dynamics and ensure more stable convergence. The main contributions of this study are as follows: 
\begin{itemize}
\item {\bf Novel transformer-based architecture:} RL-MolGAN introduces a novel Transformer-based framework that utilizes a first-decoder-then-encoder structure, effectively addressing the challenges posed by SMILES molecular representations.

\item {\bf Versatile molecular design framework:} RL-MolGAN enables both $de~novo$ and scaffold-based molecular generation, offering enhanced adaptability and flexibility in designing diverse molecular structures.

\item {\bf Property-optimized molecular generation:} By incorporating RL, RL-MolGAN not only ensures the generation of chemically valid SMILES strings but also optimizes molecules to meet specific desired chemical properties. 
\end{itemize}

The rest of the paper is organized as follows. In Section \ref{sec:related}, we review the survey of related studies. Section \ref{sec:model} introduces the RL-MolGAN framework, detailing the model architecture, including the first-decoder-then-encoder structure and the integration of RL with MCTS. In Section \ref{sec:exp}, we describe the experimental setup, datasets used, evaluation metrics, and present the results for both $de~novo$ and scaffold-based molecular generation tasks. Finally, Section \ref{sec:conclusion} concludes the paper and discusses future directions for research.

\section{Related Works}
\label{sec:related}
\subsection{De Novo Molecular Generation}
\label{subsec:denovo}
$De~novo$ molecular generation is a key area in AI-driven molecular design, focusing on the autonomous creation of novel molecules from scratch, specifically tailored to exhibit desired chemical properties \cite{sellwood2018artificial}. Recent advances have been propelled by deep learning models, such as VAEs, GANs, and diffusion models, applied to molecular representations such as SMILES, SELFIES, and graphic structures.

DeepSMILES \cite{o2018deepsmiles} employed recurrent neural networks (RNNs) \cite{graves2012long} to generate molecules from SMILES strings, successfully producing syntactically valid molecules but often struggling with long-range dependencies. This limitation led to the development of Transformer-based models such as TenGAN \cite{li2024tengan} and Chemformer \cite{irwin2022chemformer}, which more effectively capture global sequence information. Although transformers have improved the ability to preserve structural coherence in generated molecules, they still face challenges in generating valid SMILES strings due to the absence of explicit chemical constraints within the format \cite{li2022transformer}.

SELFIES addresses some of SMILES' limitations by using a self-referential format that guarantees syntactic validity. SELFIES-RNN \cite{flam2022language} generates 100\% chemically valid molecules, making it a reliable tool to explore the chemical space. However, SELFIES has certain limitations, including its complexity and low interpretability, especially when dealing with advanced molecular grammar \cite{wu2024tsis}. This complexity can make it challenging to decipher certain sequences, whereas SMILES benefits from being more widely adopted, concise, and easier to read. Furthermore, unlike SMILES, which encodes the semantics of molecules, SELFIES focuses primarily on generating syntactically valid molecular structures \cite{mukherjee2021predicting}. This distinction means that even small changes in syntax can lead to significant variations in the generated molecules' chemical composition, functional groups, and properties \cite{das2024advances}. Furthermore, SELFIES struggles to fully represent large macromolecules and crystals, particularly those with complex bonding patterns, further limiting its applicability in certain domains \cite{krenn2022selfies}.

Graph-based de novo molecular generation, where molecules are represented as graphs with atoms as nodes and bonds as edges, has gained popularity for its ability to accurately capture complex chemical structures. Models like MolGAN \cite{de2018molgan} and GraphVAE \cite{simonovsky2018graphvae} use graph neural networks to generate valid molecular graphs. MiDi \cite{vignac2023midi} and MUDiff \cite{hua2024mudiff} combine diffusion models to jointly generate molecular graphs and their corresponding 3D atom arrangements. While MiDi focuses on generating molecular structures, MUDiff extends this by integrating both discrete and continuous diffusion processes to produce a comprehensive representation, including atom features, 2D molecular structures, and 3D molecular coordinates.

\subsection{Scaffold-based Molecular Generation}
\label{subsec:scaffold}
Scaffold-based molecular generation is a key technique in molecular design, where an existing core structure, or scaffold, is modified by adding or altering functional groups to create molecules with desired properties \cite{zhao2010analoging}. This approach is particularly useful in fields like drug discovery, where the core structure provides stability and familiarity, while the functional groups introduce the diversity needed for tailoring specific biological or chemical activities \cite{jameel2018minimalist}. Modifying the scaffold allows researchers to explore a vast chemical space, generating novel molecules optimized for specific functions while preserving the core structure \cite{langevin2020scaffold}.

Several advanced models have emerged to automate scaffold-based molecular generation, each with its own unique approach to modifying scaffolds and ensuring the generation of valid, diverse molecules. For instance, SpotGAN \cite{li2023spotgan} employs a discrete GAN model combined with RL to generate molecules from a given SMILES-based scaffold. This model effectively creates new molecules by focusing on the scaffold and exploring functional group modifications while maintaining structural coherence. Sc2Mol \cite{liao2023sc2mol} starts by generating a basic scaffold made up of carbon atoms and single bonds using a VAE. This initial scaffold is sampled from a distribution of possible scaffolds, providing a broad range of novel structures. Once generated, the scaffold is modified with a transformer that alters atom and bond types, ultimately producing SMILES strings of molecules with desired properties.

SELF-EdiT \cite{piao2023self} presents an innovative approach to scaffold-based molecular generation. By iteratively applying fragment-based deletion and insertion operations to SELFIES strings, this model preserves the essential scaffolds of the original molecule while introducing modifications that result in novel structures. Its fragment-based method ensures the retention of key structural elements, such as scaffolds, while enabling the generation of diverse compounds with targeted properties. Similarly, DeepScaffold \cite{li2019deepscaffold} employs graph convolutional networks to modify scaffolds by adding functional groups, while ScaffoldVAE \cite{lim2020scaffold} uses a VAE framework to optimize both the scaffold structure and the diversity of functional group variations within molecular graphs. They utilize graph-based representations to capture the relationships between atoms, making them highly effective for scaffold-based molecular generation.

In contrast to most studies that focus solely on one generation task (either de novo or scaffold-based molecular generation), we introduce RL-MolGAN, a novel Transformer-based discrete GAN framework that combines RL and MCTS with a first-decoder-then-encoder Transformer GAN. This model is capable of generating molecules both from scratch (de novo) and based on scaffolds. Furthermore, we enhance training stability by introducing RL-MolWGAN, an extension of RL-MolGAN, which incorporates Wasserstein distance and mini-batch discrimination to improve convergence and increase diversity in the generated molecular structures. This comprehensive approach aims to advance the generation of diverse, high-quality molecules, pushing the boundaries of both de novo and scaffold-based molecular generation.

\section{Model}
\label{sec:model}
\subsection{SMILES Diversification Technique}
\label{subsec:diversified}
In $de~novo$ molecular generation, SMILES strings are typically generated by traversing the atoms of a molecule according to standardized rules, producing unique canonical forms \cite{arus2019randomized}. However, the same molecule can be represented by multiple SMILES variants depending on the traversal order of the molecular graph, known as \textit{variant SMILES} \cite{li2022transformer}. For example, the canonical SMILES “c1ccc(CC(N)=O)cc1" can appear in variant forms such as “c1c(CC(N)=O)cccc1" and “c1cccc(CC(N)=O)c1." By incorporating these variant SMILES during model training, the generator is exposed to a broader range of syntactic and semantic features, which facilitates a more comprehensive understanding of molecular representations.

Furthermore, in scaffold-based molecular generation, the challenges posed by variations in SMILES strings during sequential group extension can be addressed through a SMILES diversification technique known as \textit{diversified SMILES} \cite{li2023spotgan}. This technique involves splitting a molecule at acyclic bonds to separate scaffold and functional groups, allowing the molecule to be represented through different scaffold-functional group pairs. For example, the canonical SMILES “c1ccc(CC(N)=O)cc1” can be decomposed into various scaffold-functional group combinations, such as “c1ccc(*)cc1" paired with “CC(N)=O" and “Cc1ccc(*)cc1" paired with “C(N)=O," where “*" represents an attachment point on the scaffold for the functional group. This approach minimizes the order variations in SMILES strings, leading to more robust model learning.

\subsection{Generator: Transformer Decoder Variant}
\label{subsec:generator}

\begin{figure*}[htbp]
\centering
\includegraphics[width=0.95\textwidth]{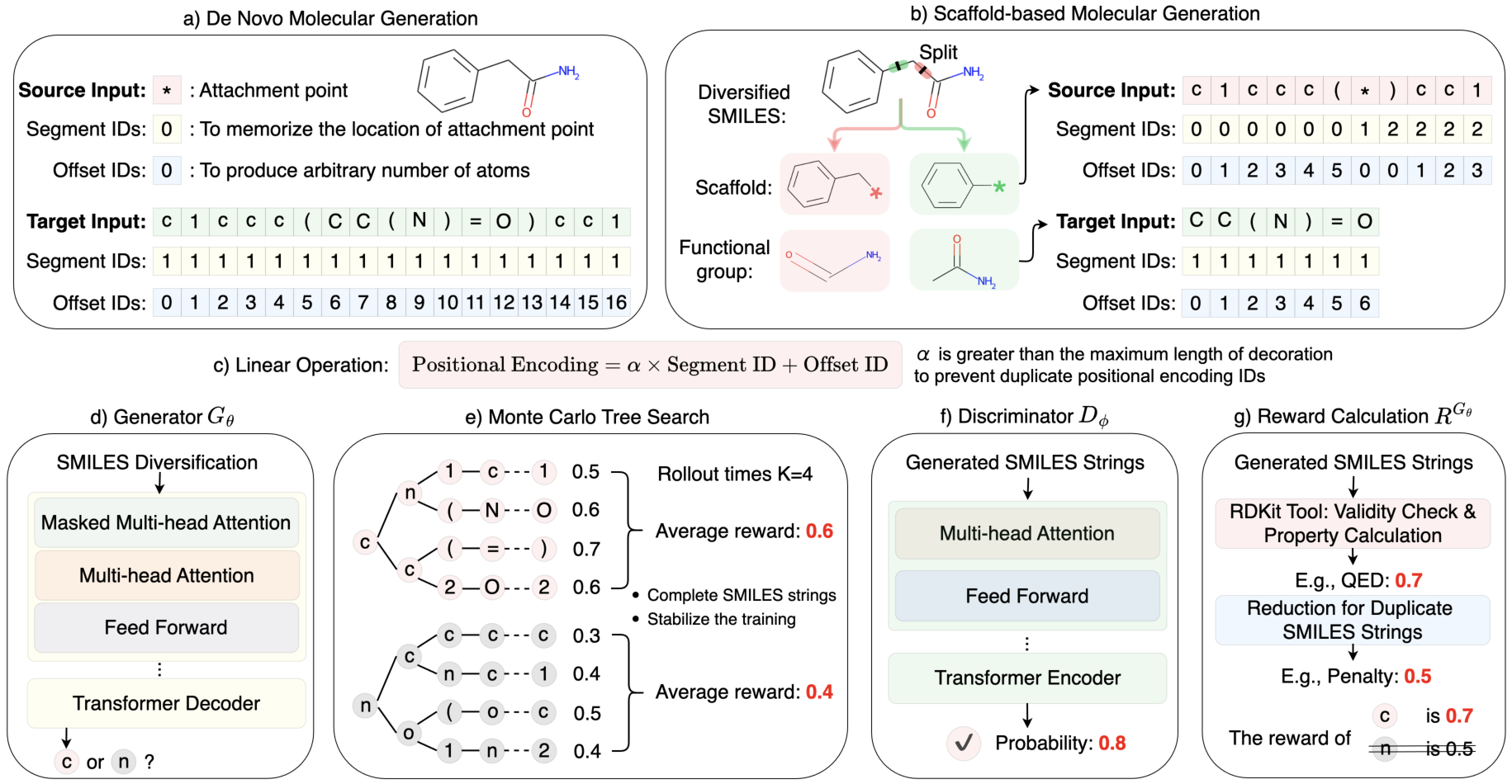}
\caption{Overview of RL-MolGAN. RL-MolGAN is an innovative molecular generation model that utilizes a transformer-based framework combining a transformer decoder followed by a transformer encoder to generate molecules for both $de~novo$ and scaffold-based tasks. {\bf a) $De~novo$ molecular generation:} the attachment points and complete SMILES representation of a molecule are used as the source and target inputs, respectively, for the transformer decoder. {\bf b) Scaffold-based molecular generation:} a given scaffold and functional group are provided as the source and target inputs, respectively, for the model. {\bf c) Position encoding:} To handle attachment point locations and generate an arbitrary number of atoms (a missing function in traditional transformers), segment IDs and offset IDs are calculated and integrated into the positional encoding through a proposed linear operation. {\bf d) Generator:} After encoding positions, a variant of the transformer decoder is used as the generator in RL-MolGAN to create molecules. {\bf e) Stabilizing training with MCTS:} To stabilize the training process of GANs and generate discrete SMILES strings while optimizing molecular properties, MCTS is applied to predict the next token in the SMILES string. {\bf f) Discriminator:} The generated SMILES-like strings are combined with the training SMILES dataset and fed into a transformer encoder-based discriminator, which evaluates whether the generated strings are real or fake. {\bf g) Reward calculation and generate update:} Finally, the output of the discriminator is combined with the calculated chemical properties and penalties for duplicated SMILES strings. The result serves as the reward signal for our RL algorithm, which is used to update the generator.}
\label{fig:rl-molgan}
\end{figure*}

In a manner similar to NLP tasks \cite{zhu2019text}, let $\bm{X}_{1:n}=[x_1, x_2, \cdots, x_n]$ and $\bm{Y}_{1:m}=[y_1, y_2, \cdots, y_m]$ represent the SMILES strings of a scaffold and a functional group, with lengths $n$ and $m$, respectively. Here, $x_i$ ($i \in \mathbb{N}_{\leq n}=\{1,\cdots, n\}$) and $y_j$ ($j \in \mathbb{N}_{\leq m}=\{1,\cdots, m\}$) denote the $i$-th and $j$-th atoms of the scaffold and functional group. The asterisk marks the attachment point where the functional group is linked to the scaffold. For example, in $\bm{X}_{1:n}=[x_1,\cdots,x_{i-1},*,x_i,\cdots,x_n]$, the functional group is attached at position $i$ within the scaffold. The complete SMILES string is then expressed as $\bm{Z}_{1:n+m}=[\bm{X}_{1:i-1},\bm{Y}_{1:m},\bm{X}_{i:n}]$, where the attachment point is replaced by the functional group, thus completing the molecular structure.

\noindent\textbf{Positional Encoding.} To design a novel GAN model for generating SMILES strings from both $de~novo$ and scaffold-based approaches, a Transformer decoder variant is utilized as the generator. Typically, a Transformer decoder requires two types of inputs: the source input and the target input. Moreover, since SMILES strings may vary in attachment positions and the number of atoms, additional positional information is incorporated. To achieve this, segment IDs and offset IDs are defined. Let $\bm{S}$ represent the segment IDs of a SMILES string, assigned based on attachment points, starting from 0. The segment IDs distinguish between scaffold and functional group within the SMILES string. For instance, given a SMILES string $\bm{Z}_{1:n+m}$, the string can be divided into three segments according to the attachment point: $\bm{S}_{\bm{X}_{1:i-1}}=[0,\cdots,0]$, $\bm{S}_{\bm{Y}_{1:m}}=[1,\cdots,1]$, and $\bm{S}_{\bm{X}_{i:n}}=[2,\cdots,2]$. Offset IDs, denoted as $\bm{O}$, are used to locate atoms within each segment and are assigned incrementally, starting from 0. For the SMILES string $\bm{Z}_{1:n+m}$, the offset IDs are defined as $\bm{O}_{\bm{X}_{1:i-1}}=[0,1,\cdots,i-2]$, $\bm{O}_{\bm{Y}_{1:m}}=[0,1,\cdots,m-1]$, and $\bm{O}_{\bm{X}_{i:n}}=[0,1,\cdots,n-i]$.

For $de~novo$ molecular generation, where no additional constraints are imposed, the model produces a complete SMILES string from scratch. In this scenario, the attachment point serves as the source input, and the full SMILES string functions as the target input for the Transformer decoder variant. To incorporate positional information, segment IDs are used to record the location of the attachment point, while offset IDs facilitate the generation of an arbitrary number of atoms. For the source input, since it consists solely of the attachment point, both segment IDs and offset IDs are set to 0. Conversely, for the target input, the segment IDs are uniformly set to 1, and the offset IDs are assigned incrementally, starting from 0 up to the maximum length of the SMILES string, as illustrated in Fig. \ref{fig:rl-molgan} a). This structured positional encoding enables the model to effectively interpret positional and contextual information, enhancing the accuracy of the generation process. Note that in $de~novo$ molecular generation, the attachment point and the complete SMILES string can be considered as a special case of the scaffold and functional group relationship.

For scaffold-based molecular generation, a given scaffold is used to guide the creation of a functional group. In this case, the scaffold serves as the source input, and the functional group serves as the target input. As illustrated in Fig. \ref{fig:rl-molgan} b), the scaffold's segment IDs start at 0, incrementing by 1 at each attachment point. The scaffold’s offset IDs start at 0, incrementing for each atom but reset to 0 after encountering an attachment point. For the target input, segment and offset IDs are assigned similarly to those in $de~novo$ molecular generation, with segment IDs uniformly set to 1 and offset IDs assigned sequentially from 0 up to the maximum length of the functional group.

Finally, the positional encoding is determined using the linear function $pos = \alpha \cdot \bm{S} + \bm{O}$, where $a \in \mathbb{N}{\leq \vert \bm{O}{\text{max}}\vert}$ and $\vert \bm{O}_{\text{max}} \vert$ represents the maximum length of the functional group. The resulting positions are processed through a sinusoidal encoding function, combined with the embeddings, and then fed into the Transformer decoder as input.

\subsection{Discriminator: Transformer Encoder Variant}
\label{subsec:discriminator}
The self-attention mechanism's capacity to manage both local and global attention fields enables each atom in a SMILES string to interact with all other atoms within the sequence. In this setup, the Transformer encoder functions as a classifier, refining the generator's outputs to enhance their drug-likeness. Unlike the generator, which utilizes scaffold and functional group SMILES representations as inputs to the Transformer decoder, the discriminator directly evaluates complete SMILES strings. As multi-head self-attention operates without masking, segment and offset IDs are unnecessary for positional computations. For a SMILES string $\bm{Z}_{1:n+m}$, positional indices are simply assigned as $pos=[0,1,\cdots,n+m-1]$.

While RL enables GANs to generate molecules from discrete SMILES strings, it introduces challenges such as unstable convergence and mode collapse. Additionally, the diversity of generated molecules tends to decline during training. This occurs because the discriminator assesses each SMILES string in isolation, disregarding correlations in the gradients. Consequently, molecules deemed valid and structurally similar are rewarded more frequently, which limits the variability in the generated molecules.

\noindent\textbf{Mini-batch Discrimination.} To mitigate the challenges mentioned above, mini-batch discrimination \cite{salimans2016improved} is employed. This approach, using nonparametric mini-batch discrimination, helps reduce the impact of hyperparameters. Let $\bm{H} \in \mathbb{R}^{B \times d_{\text{model}}}$ represent the discriminator's output before the final fully connected layer, where $B$ is the mini-batch size. The standard deviation of $\bm{H}$ is computed along the $d_{\text{model}}$ dimension for the SMILES strings in the mini-batch. This average deviation is then concatenated to $\bm{H}$ and passed through the fully connected layer. For simplicity, the mini-batch size is assumed to equal the batch size.

\noindent\textbf{Wasserstein GAN (WGAN).} To further stabilize the training process and mitigate mode collapse issues arising from the discrete nature of SMILES strings, an extended GAN variant called SpotWGAN is introduced. This model employs the Wasserstein distance to measure the difference between real and generated SMILES strings. The objective function for the discriminator is defined using the Kantorovich-Rubinstein duality, as follows:
\begin{equation}
\label{eq:wgan}
\begin{aligned}
W(\mathcal{D}_r, \mathcal{D}_z) &= \sup_{||D_{\phi}||_L \leq 1}\mathbb{E}_{\bm{Z} \sim \mathcal{D}_r(\bm{Z})}\left[D_{\phi}(\bm{Z})\right]-\mathbb{E}_{\bm{Z} \sim \mathcal{D}_z}\left[D_{\phi}(G_{\theta}(\bm{Z}))\right],
\end{aligned}
\end{equation}
where $\mathcal{D}_r$ and $\mathcal{D}z$ represent the distributions of real and generated SMILES strings, respectively. The parameter $\sup$ denotes the least upper bound, while $D{\phi}$ is a 1-Lipschitz function constrained by this condition.

\subsection{Reward Calculation}
\label{subsec:reward}
\noindent\textbf{Reinforcement Learning.} RL, particularly the policy gradient approach \cite{sutton1999policy}, helps optimize the policy ($G_{\theta}$) while incorporating chemical properties as rewards throughout the training process. The generator aims to maximize the expected reward score, which is expressed as:
\begin{equation}
J(\theta) = \sum_{y_j\in \bm{Y}_{1:m}}G_{\theta}(y_j \vert \bm{X}_{1:n}, \bm{Y}_{1:j-1}) R^{G_{\theta}}(\bm{Y}_{1:j-1}, y_j),
\end{equation}
where $G_{\theta}$ represents the generator’s policy model parameterized by $\theta$, and $R^{G_{\theta}}$ is the action-value function that quantifies the average reward for the chemical properties of the state $\bm{Y}_{1:j-1}$, taking action $y_j$ according to the policy $G{\theta}$. For a functional group $\bm{Y}_{1:m}$, $R^{G{\theta}}$ is computed as:
\begin{equation}
\small
R^{G_{\theta}}(\bm{Y}_{1:m-1}, y_m)=\lambda D_{\phi}(\bm{Z}_{1:n+m})+(1-\lambda)\left[R(\bm{Z}_{1:n+m})P(\bm{Z}_{1:n+m})-b(\bm{Z}_{1:n+m})\right],
\end{equation}
where $\lambda \in [0, 1]$ controls the balance between RL and GAN components. The term $R$ denotes the property score, which can be computed using RDKit \cite{landrum2013rdkit}, while $P$ represents a penalty for generating duplicate SMILES strings, defined as:
\begin{equation}
P(\bm{Z}_{1:n+m}) = \frac{\# \text{~unique SMILES}}{(\#~\text{SMILES} \times \# \text{~repeated~} \bm{Z}_{1:n+m})}.
\end{equation}
The parameter $b$ serves as the baseline, below which molecules with low chemical-property scores are penalized. For simplicity, the average property score is used for $R$.

\noindent\textbf{Monte Carlo Tree Search.} Since SMILES strings are discrete data, the generator constructs them sequentially in a stepwise regressive manner, where each atom or token is sampled only after the preceding one. This process poses two major challenges when using GANs for SMILES generation. First, feedback from the discriminator is available only after the complete SMILES string is generated, resulting in delayed evaluation. Second, the generation cannot be directly steered to ensure that the produced molecules are valid or meet specified chemical properties. 

To complete SMILES strings and stabilize the training process, MCTS is employed. Specifically, for an incomplete target input $\bm{Y}_{1:j}$,  intermediate rewards are calculated at each step by conducting $K$ Monte Carlo (MC) searches:
\begin{equation}
\bm{Z}^k_{1:n+m}=[\bm{X}_{1:i-1}, \bm{Y}^k_{1:m},\bm{X}_{i:n}],~~\bm{Y}^k_{1:m}\in\text{MC}^{G_{\theta}}(\bm{Y}_{1:j}, K),\text{~and~} k\in[1,K],
\end{equation}
where $\bm{Y}^k_{1:m}=[\bm{Y}^k_{1:j},\bm{Y}^k_{j+1:m}]$ denotes the complete functional group of the $k$-th MC search. $\bm{Y}^k_{1:j} = \bm{Y}_{1:j}$, and $\bm{Y}^k_{j+1:M}$ is sampled under policy $G_{\theta}$. $R^{G_{\theta}}$ is calculated as
\begin{equation}
\label{eq:incomplete_spotgan}
\begin{aligned}
R^{G_{\theta}}(\bm{Y}_{1:j-1}, y_j)=&\frac{1}{K}\sum_{k=1}^K \lambda D_{\phi}(\bm{Z}^k_{1:n+m})+ \\
&(1-\lambda)\left[R(\bm{Z}^k_{1:n+m})P(\bm{Z}^k_{1:n+m})-b(\bm{Z}^k_{1:n+m})\right].
\end{aligned}
\end{equation}
The gradient of the objective function is calculated as
\begin{equation}
\label{eq:train_generator}
J(\theta) \simeq \frac{1}{m}\sum_{j=1}^m\sum_{y_j}R^{G_{\theta}}(\bm{Y}_{1:j-1},y_j)\nabla\log G_{\theta}(y_j\|\bm{X}_{1:n},\bm{Y}_{1:j-1}).
\end{equation}
Finally, the policy is updated by adjusting the generator parameters: $ \theta \leftarrow \theta + \nabla J(\theta)$.

Algorithm \ref{alg:algorithm} provides an overview of RL-MolGAN. Initially, the generator $G_{\theta}$ is pre-trained using maximum likelihood estimation on the real SMILES dataset $\mathcal{D}_r$. Afterward, the generator generates a generated dataset $\mathcal{D}_z$, containing the same number of SMILES strings as $\mathcal{D}r$, ensuring a balanced dataset. These datasets are then shuffled before pre-training the discriminator $D{\phi}$. Finally, adversarial training alternates between updating the generator and discriminator, with the generator’s parameters $\theta$ being refined using MCTS and policy gradient updates.
\begin{algorithm}[t]
\caption{Procedures of RL-MolGAN}
\label{alg:algorithm}
\begin{algorithmic}[1]

\State \textbf{Input:} A SMILES dataset $\mathcal{D}_r$, 
\State \textbf{Initialization:} Generator $G_{\theta}$, Discriminator $D_{\phi}$

\Statex \hspace{0em}\textit{//Pre-process the dataset $\mathcal{D}_r$.}
\State Variant SMILES or diversified SMILES is used to enhance SMILES diversification.

\Statex \hspace{0em}\textit{//Pre-train the generator on $\mathcal{D}_r$.} 
\For{$i = 1$ to $f\_epochs$}
\State Update $\theta$ with maximum likelihood estimation.
\EndFor
\State Generate a dataset $\mathcal{D}_z$ 
\State Shuffle datasets $\mathcal{D}_r$ and $\mathcal{D}_z$.
    
\Statex \hspace{0em}\textit{//Pre-train the discriminator on the shuffled dataset.}
\For{$i = 1$ to $d\_epochs$}
\State Update $\phi$ with cross-entropy or Wasserstein distance.
\EndFor
    
\Statex \hspace{0em}\textit{//Adversarial training of $G_{\theta}$ and $D_{\phi}$.} 
\For{$i = 1$ to $epochs$}

\Statex \hspace{0em}\textit{//Train the generator $G_{\theta}$}
\For{$j = 1$ to $f\_steps$}
\State Generate a dataset $\mathcal{D}_z$ 
\State Shuffle it with $\mathcal{D}_r$.

\Statex \hspace{0em}\textit{//Check if RL-MolGAN is executed.}    
\If{RL-MolGAN is executed}
\State Compute $R^{G_{\theta}}$.
\EndIf

\Statex \hspace{0em}\textit{//Check if RL-MolWGAN is executed.} 
\If{RL-MolWGAN is executed}
\State Execute RL-MolWGAN.
\EndIf
\State Update $\theta$.
\EndFor
        
\Statex \hspace{0em}\textit{//Train the discriminator $D_{\phi}$}
\For{$k = 1$ to $d\_steps$}
\State Update discriminator's parameters of $\phi$.
\EndFor
\EndFor
\end{algorithmic}
\end{algorithm}

\section{Experiments}
\label{sec:exp}
\noindent\textbf{Datasets.} Two subsets of 10,000 SMILES strings were randomly selected from the QM9 \cite{ramakrishnan2014quantum} and ZINC \cite{irwin2012zinc} datasets for training. Each SMILES string was constrained to a maximum of nine heavy atoms, including carbon (C), oxygen (O), nitrogen (N), and fluorine (F). During the preprocessing stage for scaffold-based molecular generation, each SMILES string was split into scaffold and functional group pairs. Due to the greater complexity of SMILES strings in the ZINC dataset compared to QM9, the average length of the functional groups in ZINC was approximately four times longer than in QM9.

\noindent\textbf{Implementation Details.} For both $de~novo$ and scaffold-based molecular generation, the RL-MolGAN generator consisted of four decoder layers, each equipped with four attention heads. The attention heads had a dimensionality of 128 for the QM9 dataset and 256 for the ZINC dataset. A feedforward layer of 100 dimensions was employed for both datasets. Pre-training of the generator was conducted for 100 epochs on QM9 and 200 epochs on ZINC, using a learning rate of $1\mathrm{e}{-5}$. During sampling, 10,000 molecules were generated per epoch, with maximum lengths of $30$ and $69$ for $de~novo$ generation, and $20$ and $50$ for scaffold-based generation, on QM9 and ZINC, respectively. The batch size was set to $64$, and the dropout probability was fixed at $0.1$.

The discriminator consisted of four encoder layers, each featuring four attention heads and 128-dimensional representations. The feedforward layer was configured with 200 dimensions. For RL-MolGAN and RL-MolWGAN, the learning rates were set to $1\mathrm{e}{-4}$ and $1\mathrm{e}{-5}$, respectively. The discriminator underwent pre-training for 10 epochs before adversarial training. During scaffold-based molecular generation, the generator’s learning rate in adversarial training was $2\mathrm{e}{-5}$, with $\lambda$ and $K$ parameters set to 0.5 and 8, respectively. Training was conducted for up to 100 epochs on the QM9 dataset and 50 epochs on the ZINC dataset. All experiments were executed on an NVIDIA GV100GL GPU environment.

\subsection{Evaluation Metrics}
\label{subsec:metrics}
\noindent\textbf{Statistical Measures.} The performance of the model was evaluated using several key metrics, each designed to assess a different aspect of the generated molecules. These include:
\begin{itemize}
\item {\bf Validity} refers to the proportion of chemically valid molecules among all generated molecules. This was determined using the RDKit tool in practice.
\item {\bf Uniqueness} (abbreviated as \textit{“unique"}) measures the proportion of non-duplicated molecules among all valid molecules.
\item {\bf Novelty} is the proportion of unique molecules that are not present in the training set.
\end{itemize}

\noindent{\bf Property Optimization.} The following characteristics were optimized in the model to ensure the generation of molecules with desirable chemical properties:

\begin{itemize}
\item {\bf Drug-likeness} indicates how similar a molecule is to known drugs and is quantified using the quantitative estimate of drug-likeness (QED) \cite{bickerton2012quantifying}. A higher QED score suggests that the molecule is more likely to be pharmacologically relevant.
\item {\bf Solubility}: Solubility reflects the lipophilicity (hydrophobicity) of a molecule, which is critical for drug absorption. It is measured by the octanol-water partition coefficient (logP) \cite{comer2001lipophilicity}. Molecules with a balanced logP value are more likely to have good solubility properties.
\item {\bf Synthesizability}: This property measures how easily a molecule can be synthesized in the lab. It is quantified using the synthetic accessibility (SA) score \cite{ertl2009estimation}. A higher SA score indicates that the molecule is easier to synthesize, which is important for practical drug development.
\end{itemize}

For consistency in the reward system, all of these measures range from 0 (worst) to 1 (best) and were used to guide the model’s training process. 

\begin{table}[t]
\centering
\caption{Statistical results comparing RL-MolGAN with baselines for $de~novo$ and scaffold-based molecular generation on two datasets.}
\label{tab:res1}
\resizebox{\columnwidth}{!}{%
\begin{threeparttable}
\begin{tabular}{cclcccc}\toprule
Goal&Pattern&Model&Validity (\%)&Unique (\%)&Novelty (\%)&Total (\%)\\ \toprule 
\multirow{10}{12em}[-8pt]{$De~novo$-based (QM9)}
&\multirow{5}{6em}[0pt]{Graph-based}
&JTVAE \cite{jin2018junction}&\colorbox[gray]{0.9}{100}&55.7&97.1&54.1\\
&&GraphAF \cite{shi2020graphaf}&37.0&91.7&99.6&33.8\\
&&GraphNVP \cite{madhawa2019graphnvp}&83.0&99.2&-&-\\
&&MolGAN \cite{de2018molgan}&99.3&2.3&\colorbox[gray]{0.9}{99.7}&2.3\\
&&MoFlow \cite{zang2020moflow}&95.0&93.7&89.0&\colorbox[gray]{0.9}{79.2}\\\cmidrule(r){2-7}
&\multirow{5}{7em}[0pt]{SMILES-based}
&ORGAN \cite{guimaraes2017objective}&88.1&65.7&97.9&56.7\\
&&CharVAE \cite{kusner2017grammar}&17.2&\colorbox[gray]{0.9}{99.9}&94.9&16.3\\
&&GramVAE \cite{kusner2017grammar}&38.0&98.8&93.7&35.2\\ 
&&TransVAE \cite{dollar2021giving}&17.2&25.2&97.2&42.1\\
&&{\bf RL-MolGAN} &84.8&89.3&99.6&75.4\\\toprule

\multirow{6}{12em}[0pt]{Scaffold-based (ZINC)}
&Motifs-based&MoLeR \cite{maziarz2021learning}&\colorbox[gray]{0.9}{94.5}&88.2&88.2&73.5\\\cmidrule(r){2-7}
&\multirow{4}{7em}[0pt]{SMILES-based}
&Na\"{\i}ve RL \cite{guimaraes2017objective}&89.2&88.5&91.9&72.5\\
&&SCRNN \cite{langevin2020scaffold}&88.7&86.6&86.6&66.5\\
&&SCRNN-RL \cite{langevin2020scaffold}&79.3&\colorbox[gray]{0.9}{93.7}&\colorbox[gray]{0.9}{94.7}&70.4\\
&&{\bf RL-MolGAN}&93.3&92.8&92.8&\colorbox[gray]{0.9}{80.4}\\\toprule
\end{tabular}
\begin{tablenotes}
\footnotesize
\item[$\star$] “Total" is the product of validity, uniqueness, and novelty. The values shaded in gray correspond to the maximum values.
\end{tablenotes}
\end{threeparttable}
}
\end{table}

\subsection{Research Questions}
\label{subsec:rqs}
The experiments were conducted to address the following research questions (RQs):
\begin{itemize}
\item {\bf RQ1:} How does RL-MolGAN perform according to statistical measures?
\item {\bf RQ2:} How effective is RL-MolGAN in molecular optimization?
\item {\bf RQ3:} What are the effects of different diversification techniques, WGAN, and key hyperparameters?
\item {\bf RQ4:} Is bioactivity optimization for a therapeutic target effective?
\end{itemize}

\subsection{RQ1: Evaluation of Statistical Measures}
\label{subsec:eval_1}
To address RQ1, we present the statistical measure results comparing RL-MolGAN with baseline models in Table \ref{tab:res1}. For the evaluation of $de~novo$ molecular generation, we compared both graph-based models (JTVAE, GraphAF, GraphNVP, MolGAN, and MoFlow) and SMILES-based models (ORGAN, CharVAE, GramVAE, and TransVAE) on the QM9 dataset. Our results reveal that MoFlow and RL-MolGAN achieved the best and second-best Total scores (79.2\% and 75.4\%, respectively) compared to all other methods. This highlights the higher performance of graph-based models, which typically contain more structural information than SMILES strings. Nevertheless, RL-MolGAN still outperforms other SMILES-based models, demonstrating its effectiveness in $de~novo$ molecular generation.

Additionally, for scaffold-based molecular generation, we compared the performance of RL-MolGAN with several state-of-the-art (SOTA) models, including motifs-based MoLeR and SMILES-based Na\"{\i}ve RL, SCRNN, and SCRNN-RL. The results indicate that RL-MolGAN achieved the highest performance with a total score of 80.4\%, outperforming all other baselines.

Overall, RL-MolGAN exhibits significant effectiveness and high performance in both $de~novo$ molecular generation and scaffold-based molecular generation tasks.

\subsection{RQ2: Performance of Property Optimization}
\label{sec:property}
\begin{table}[t]
\setlength\tabcolsep{6pt}
\centering
\caption{Average chemical property scores of Top-1000 molecules generated by RL-MolGAN and baseline models.}
\label{tab:top_1000_res}
\begin{tabular}{clccc}\toprule
Goal & Model & QED & logP & SA \\\cmidrule(r){1-5}
\multirow{3}{7em}[0pt]{$De~novo$-based (QM9)}
&ORGAN \cite{guimaraes2017objective}&0.55&0.54&0.71 \\
&GraphNVP \cite{madhawa2019graphnvp}&0.42&-&- \\
&{\bf RL-MolGAN} &\colorbox[gray]{0.9}{0.68}&\colorbox[gray]{0.9}{0.83}&\colorbox[gray]{0.9}{0.94} \\\cmidrule(r){1-5}

\multirow{3}{7em}[0pt]{Scaffold-based (ZINC)}
&Transformer \cite{li2023spotgan}&0.47&0.30&0.33 \\
&Na\"{\i}ve RL \cite{guimaraes2017objective}&0.57&0.80&0.75\\
&{\bf RL-MolGAN}&\colorbox[gray]{0.9}{0.92}&\colorbox[gray]{0.9}{0.97}&\colorbox[gray]{0.9}{0.98} \\\toprule
\end{tabular}
\end{table}
Table \ref{tab:top_1000_res} presents the average scores of three chemical properties (QED, logP, and SA) for the top 1000 molecules generated by our RL-MolGAN model and the baseline models. For $de~novo$ molecular generation, RL-MolGAN exhibited superior performance, outperforming the ORGAN baseline model by 23.6\%, 53.7\%, and 32.4\% in QED, logP, and SA. Similarly, in scaffold-based molecular generation, RL-MolGAN surpassed the pure Na\"{\i}ve RL-based method by 61.4\%, 21.3\%, and 30.7\% for these properties.

\begin{figure*}[ht]
\centering
\begin{subfigure}{0.3\textwidth}
\centering
\includegraphics[width=\linewidth]{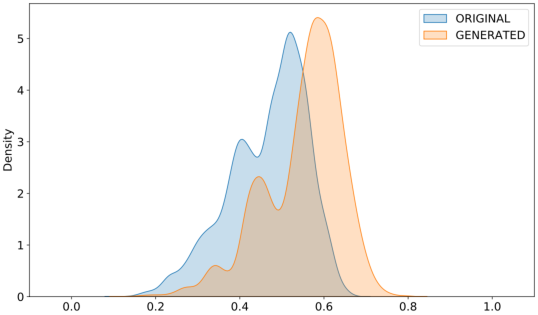}
\caption{QED distribution.}
\end{subfigure}
\hspace{-4pt}
\begin{subfigure}{0.3\textwidth}
\centering
\includegraphics[width=\linewidth]{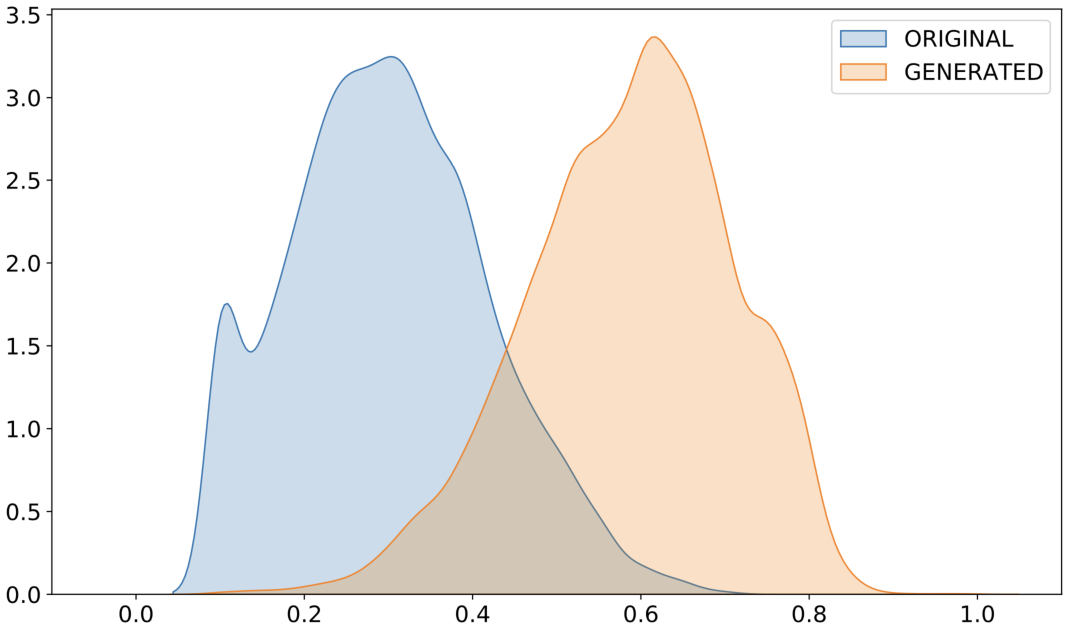}
\caption{logP distribution.}
\end{subfigure}
\hspace{-4pt}
\begin{subfigure}{0.3\textwidth}
\centering
\includegraphics[width=\linewidth]{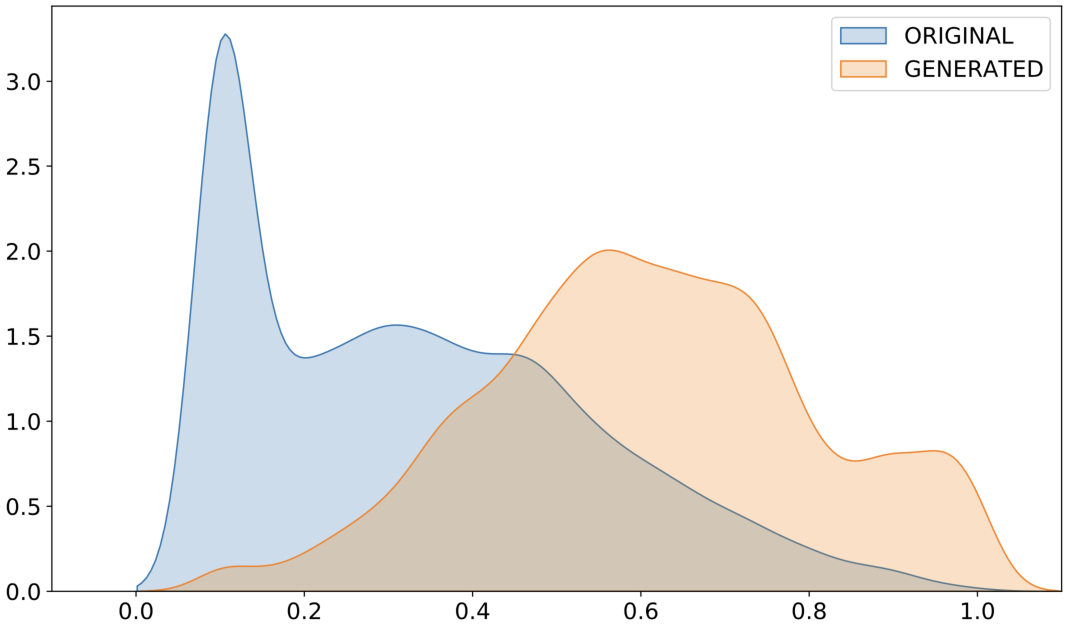}
\caption{SA distribution.}
\end{subfigure}
\caption{Property distributions of QM9 dataset and those generated by RL-MolGAN for $de~novo$ molecular generation.}
\label{fig:qm9_property}
\end{figure*}    

\begin{figure*}[ht]
\centering
\begin{subfigure}{0.3\textwidth}
\centering
\includegraphics[width=\linewidth]{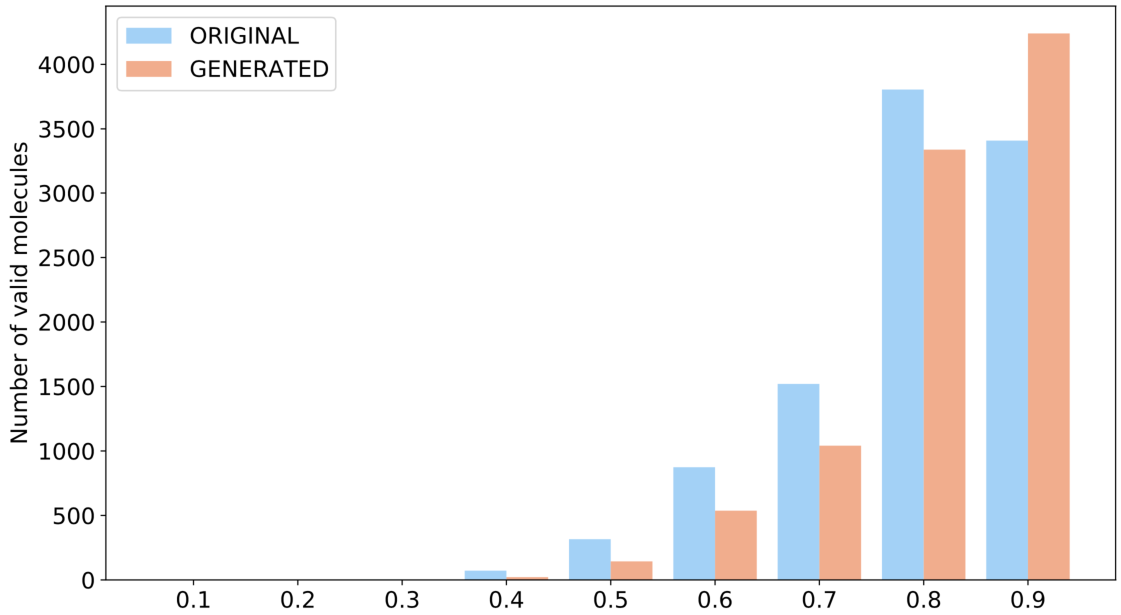}
\caption{QED distribution.}
\end{subfigure}
\hspace{-4pt}
\begin{subfigure}{0.3\textwidth}
\centering
\includegraphics[width=0.98\linewidth]{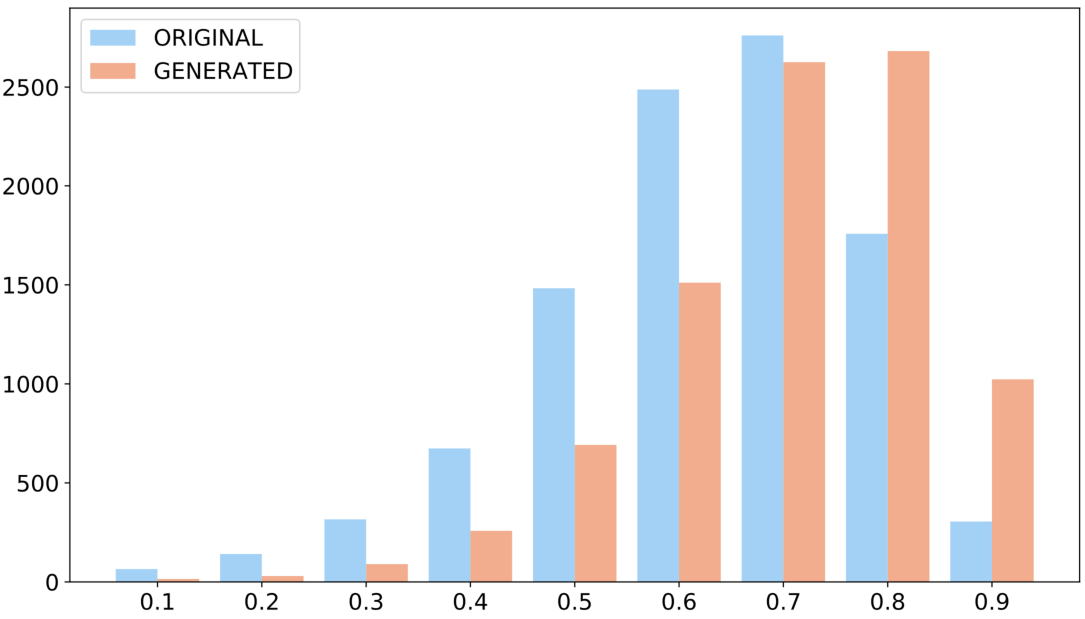}
\caption{logP distribution.}
\end{subfigure}
\hspace{-4pt}
\begin{subfigure}{0.3\textwidth}
\centering
\includegraphics[width=\linewidth]{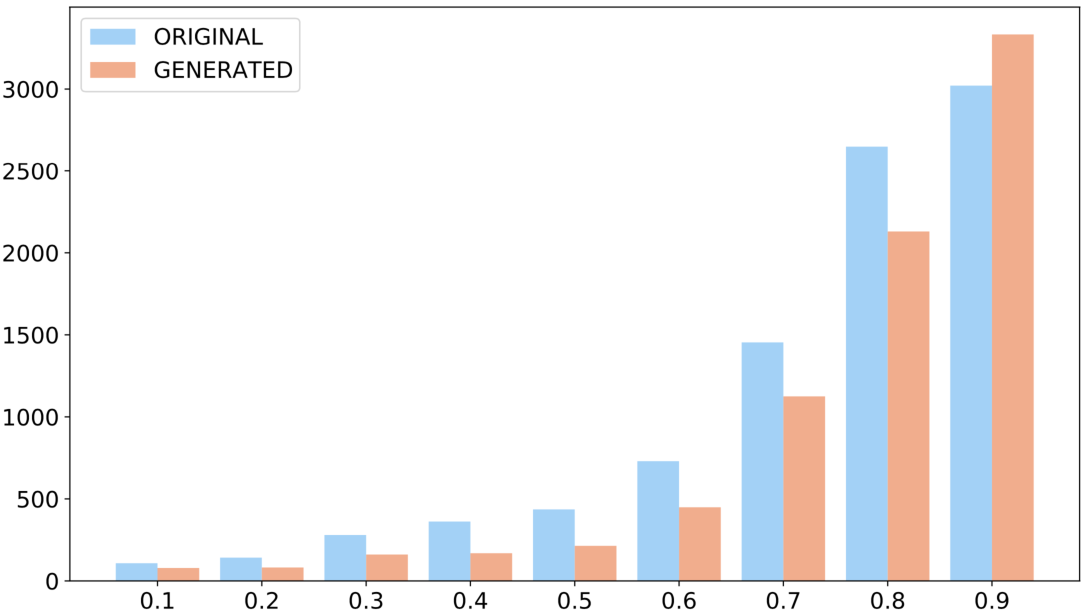}
\caption{SA distribution.}
\end{subfigure}
\caption{Property distributions of ZINC dataset and those generated by RL-MolGAN for scaffold-based molecular generation.}
\label{fig:zinc_property}
\end{figure*}
Figures \ref{fig:qm9_property} and \ref{fig:zinc_property} illustrate the distributions of chemical properties for the QM9 and ZINC datasets generated by RL-MolGAN in $de~novo$ and scaffold-based molecular generation, respectively. The results show a noticeable rightward shift in the property distributions of molecules generated by RL-MolGAN (GENERATED) compared to the training sets (ORIGINAL). This indicates that RL-MolGAN successfully generates molecules with higher chemical properties, addressing the requirements of molecular property optimization tasks.

Figures \ref{fig:qm9_change} and \ref{fig:zinc_change} in the appendix illustrate the changes in three property scores over training epochs for the RL-MolGAN model on the QM9 and ZINC datasets, respectively. The results indicate that all three property scores improved consistently during training on both datasets. Specifically, for $de~novo$ molecular generation, the absence of fixed scaffold constraints allows for greater flexibility in the generated molecules, leading to more significant increases in chemical property scores. In contrast, for scaffold-based molecular generation, despite the limitations imposed by predefined scaffolds, RL-MolGAN is still able to produce molecules with high property scores.

Figures \ref{fig:top12_denovo_qm9}-\ref{fig:zinc_sa_12} in the appendix showcase the top 12 molecular structures generated by RL-MolGAN for both $de~novo$ and scaffold-based molecular generation tasks across the two datasets. The results demonstrate that the model successfully produced novel, drug-like molecules with high property scores.

Overall, RL-MolGAN consistently outperformed other approaches in both $de~novo$ and scaffold-based molecular generation, demonstrating its robust capability in optimizing molecular properties.

\subsection{RQ3: Ablation Studies}
\label{sec:ablation}
\begin{table}[t]
\centering
\caption{Effect of SMILES diversification techniques.}
\label{table:effect_smiles}
\centering
\begin{tabular}{lcccc}\toprule
&QED&Validity (\%)&Unique (\%)&Novelty (\%)\\\toprule 
w/o&0.51&90.1&89.2&83.0\\
w/&\colorbox[gray]{0.9}{0.53}&\colorbox[gray]{0.9}{93.3}&\colorbox[gray]{0.9}{94.2}&\colorbox[gray]{0.9}{91.3}\\\toprule
\end{tabular}
\end{table}
\begin{table}[t]
\caption{Effect of WGAN on drug-likeness.}
\label{table:wgan_res}
\centering
\begin{tabular}{lccccc}\toprule
&QED&Validity (\%)&Unique (\%)&Novelty (\%)\\\toprule 
RL-MolGAN&\colorbox[gray]{0.9}{0.53}&93.3&94.2&91.3\\
RL-MolWGAN&\colorbox[gray]{0.9}{0.53}&\colorbox[gray]{0.9}{94.7}&\colorbox[gray]{0.9}{94.4}&\colorbox[gray]{0.9}{91.9}\\\toprule 
\end{tabular}
\end{table}
\begin{table}[t]
\caption{Effect of $\lambda$ on the drug-likeness of RL-MolGAN.}
\label{table:effect_lambda}
\centering
\centering
\begin{tabular}{lccccc}\toprule
$\lambda$&QED&Validity (\%)&Unique (\%)&Novelty (\%)\\ \toprule 
0&0.53&93.8&94.2&93.8\\
0.1&0.53&96.4&90.4&85.6\\
0.3&0.53&94.1&92.0&91.6\\
0.5&0.53&93.3&94.2&91.3\\
0.7&0.53&93.0&93.1&95.1\\
0.9&0.52&92.6&93.5&\colorbox[gray]{0.9}{97.1}\\
1.0&0.47&\colorbox[gray]{0.9}{96.4}&\colorbox[gray]{0.9}{94.9}&73.6\\ \toprule 
\end{tabular}
\end{table}
\begin{table}[t]
\setlength\tabcolsep{7pt}
\centering
\caption{Effect of $K$ of RL-MolGAN.}
\label{table:effect_n}
\centering
\begin{tabular}{lccccc}\toprule
$K$&Validity (\%)&Unique (\%)&Novelty (\%)&Time (h)\\ \toprule 
2&91.7&94.1&94.0&2.5\\
4&91.4&93.5&92.5&2.4\\
8&\colorbox[gray]{0.9}{93.3}&\colorbox[gray]{0.9}{94.2}&91.3&3.1\\
16&92.6&93.3&90.5&4.0\\
32&92.2&94.2&\colorbox[gray]{0.9}{95.1}&6.9\\\toprule 
\end{tabular}
\end{table}
To address RQ3, we conducted ablation studies to evaluate the effectiveness of diversification techniques, WGAN, and key hyperparameters ($\lambda$ and $K$) on the QM9 dataset.

As shown in Table \ref{table:effect_smiles}, the diversified SMILES technique significantly improved both statistical metrics and chemical property optimization. Specifically, it enhanced validity by 3.6\%, uniqueness by 5.6\%, novelty by 10.0\%, and QED by 3.9\%. These findings indicate that the proposed SMILES diversification techniques enabled RL-MolGAN to better extract meaningful features from SMILES representations, thereby improving its performance in molecular generation tasks.

Additionally, the results in Table \ref{table:wgan_res} demonstrated that incorporating RL-MolWGAN improved validity by 1.5\%, 0.2\%, and 0.7\%. These findings show WGAN's ability to stabilize GAN training, contributing to enhanced performance in molecular generation tasks.

Furthermore, Tables \ref{table:effect_lambda} and \ref{table:effect_n} illustrate the impact of two key hyperparameters, $\lambda$ and $K$, on the performance of RL-MolGAN. The hyperparameter $\lambda$ controls the trade-off between the RL and GAN objectives. We evaluated values ranging from $[0, 0.1, 0.3, 0.5, 0.7, 0.9, 1.0]$, where $\lambda=0$ indicates exclusive dependence on RL, and $\lambda=1.0$ indicates exclusive reliance on GAN. The results show that adjusting $\lambda$ effectively improved property scores, underscoring its role in balancing learning objectives.

\begin{figure}[t]
\centering
\includegraphics[width=0.96\textwidth]{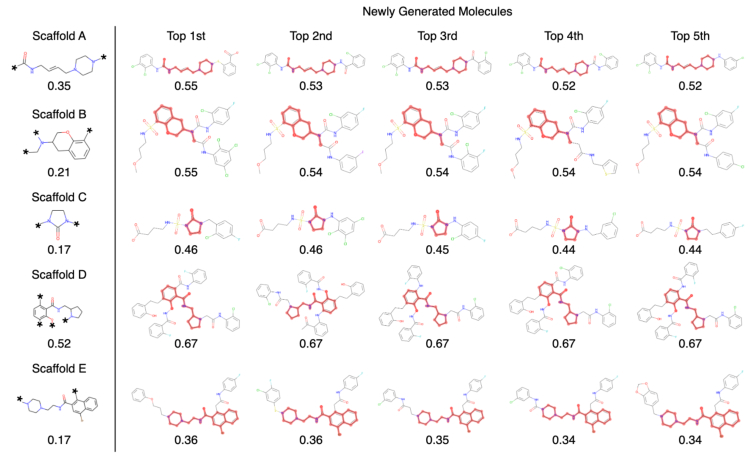}
\caption{top-5 molecular structures for scaffolds A–E along with their corresponding DRD2 scores on the ZINC dataset.}
\label{fig:top_scaffolds}
\end{figure}

The parameter $K$ represents the number of rollout iterations in the MCTS. We evaluated the values of $K \in [2, 4, 8, 16, 32]$. Although increasing $K$ generally improved performance, it also significantly increased training time, with approximate time costs of 2.5 hours at $K=2$ and 6.9 hours at $K=32$. These findings show the trade-off between computational cost and model performance when tuning $K$.

\subsection{RQ4: Case Studies}
\label{sec:case}
To answer RQ4, we also conducted case studies to explore bioactivity optimization for the therapeutic target dopamine receptor D2 (DRD2) \cite{olivecrona2017molecular} using the ZINC dataset. A higher DRD2 score indicates a greater chance that the molecule has therapeutic potential for the specified target. The results are presented in Fig. \ref{fig:top_scaffolds}. The five widely used scaffolds, A, B, C, D, and E, were chosen as the attachment points for molecule generation, given their known efficacy for DRD2 ligands. These scaffolds were modified with 2, 3, 2, 4, and 2 attachment points, as indicated by the asterisks on the left side of the figure. The top five generated molecules and their respective DRD2 scores are displayed on the right side of the figure. The DRD2 scores of the newly generated molecules are notably higher than those of the original scaffolds, a pattern observed across all five scaffolds. Interestingly, all the generated molecules adhered to H{\"u}ckel's rules \cite{klein1984huckel}, which are essential for achieving the desired chemical properties in drug development. These results suggest that RL-MolGAN successfully generated drug-like molecules with enhanced DRD2 scores for scaffold-based molecular generation.

\section{Conclusion}
\label{sec:conclusion}
RL-MolGAN marks a significant advancement in the field of molecular generation, providing a robust solution to the challenges of designing molecules with specific chemical properties. By introducing a novel Transformer-based discrete GAN framework with a first-decoder-then-encoder architecture, RL-MolGAN is capable of generating structurally valid, drug-like molecules. The integration of RL and MCTS ensures stability during training while optimizing the chemical properties of the generated molecules. Additionally, the extended version, RL-MolWGAN, further enhances training stability through advanced techniques such as Wasserstein distance and mini-batch discrimination.

Our approach offers flexibility for both $de~novo$ and scaffold-based molecular generation, enabling the design of molecules with tailored properties. Experimental results on benchmark datasets, QM9 and ZINC, demonstrate the effectiveness of RL-MolGAN in generating high-quality molecular structures. This work sets the stage for further advancements in AI-driven molecular design, paving the way for more efficient and accelerated drug discovery and chemical synthesis.

\section*{Acknowledgments}
This research has received no external funding.

\bibliography{refs}
\clearpage
\setcounter{secnumdepth}{2}
\numberwithin{figure}{section}
\numberwithin{table}{section}
\appendix
\renewcommand{\thesection}{\Alph{section}}
\renewcommand{\thefigure}{A.\arabic{figure}}
\setcounter{section}{0}
\setcounter{figure}{0}
\setcounter{table}{0}
\section{Details of Experimental Results}
\label{app:exp}
\begin{figure}[ht]
\centering
\begin{subfigure}{0.42\textwidth}
\centering
\includegraphics[width=\textwidth]{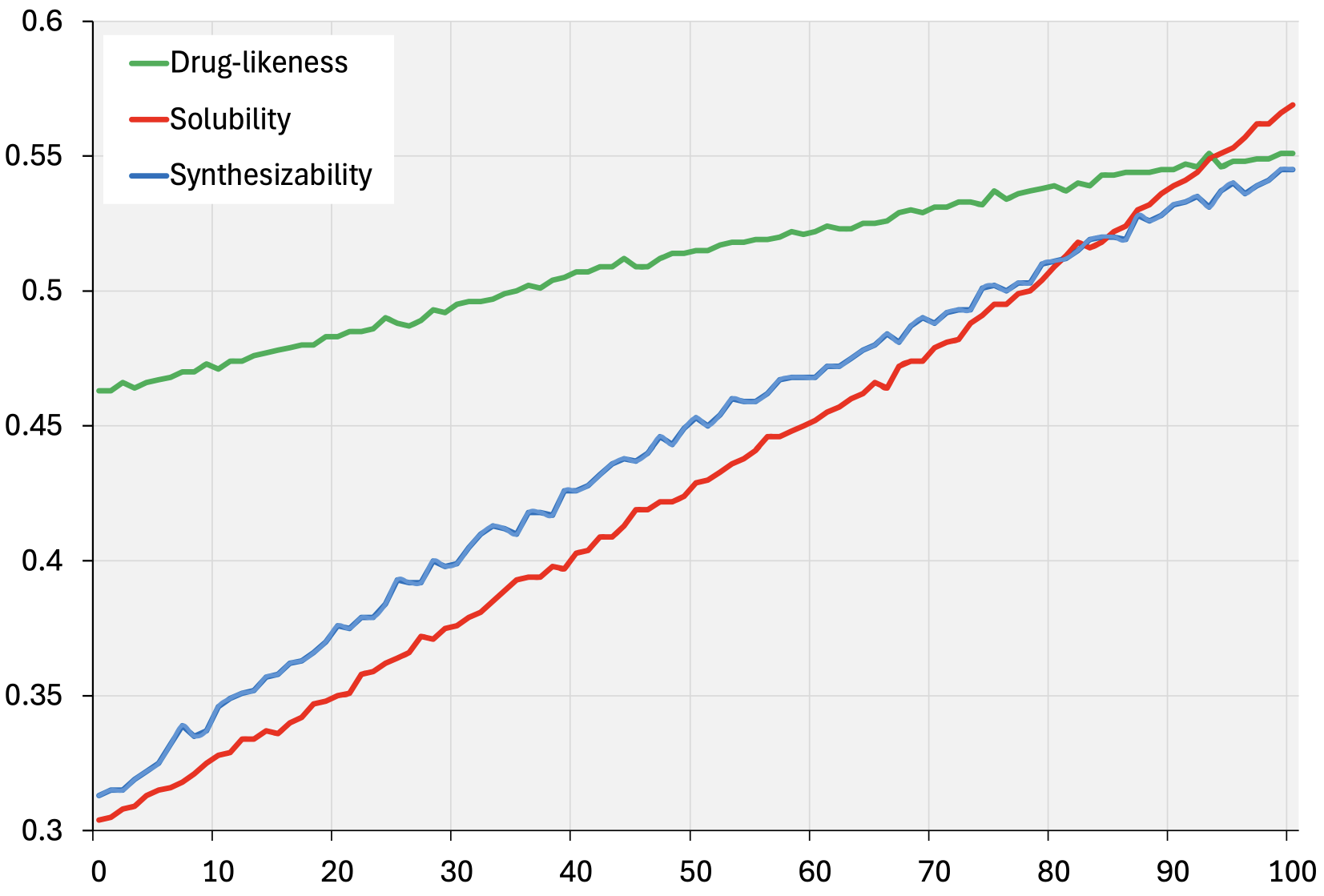}
\caption{$De~novo$ molecular generation.}
\end{subfigure}
\hspace{-4pt}
\begin{subfigure}{0.5\textwidth}
\centering
\includegraphics[width=\textwidth]{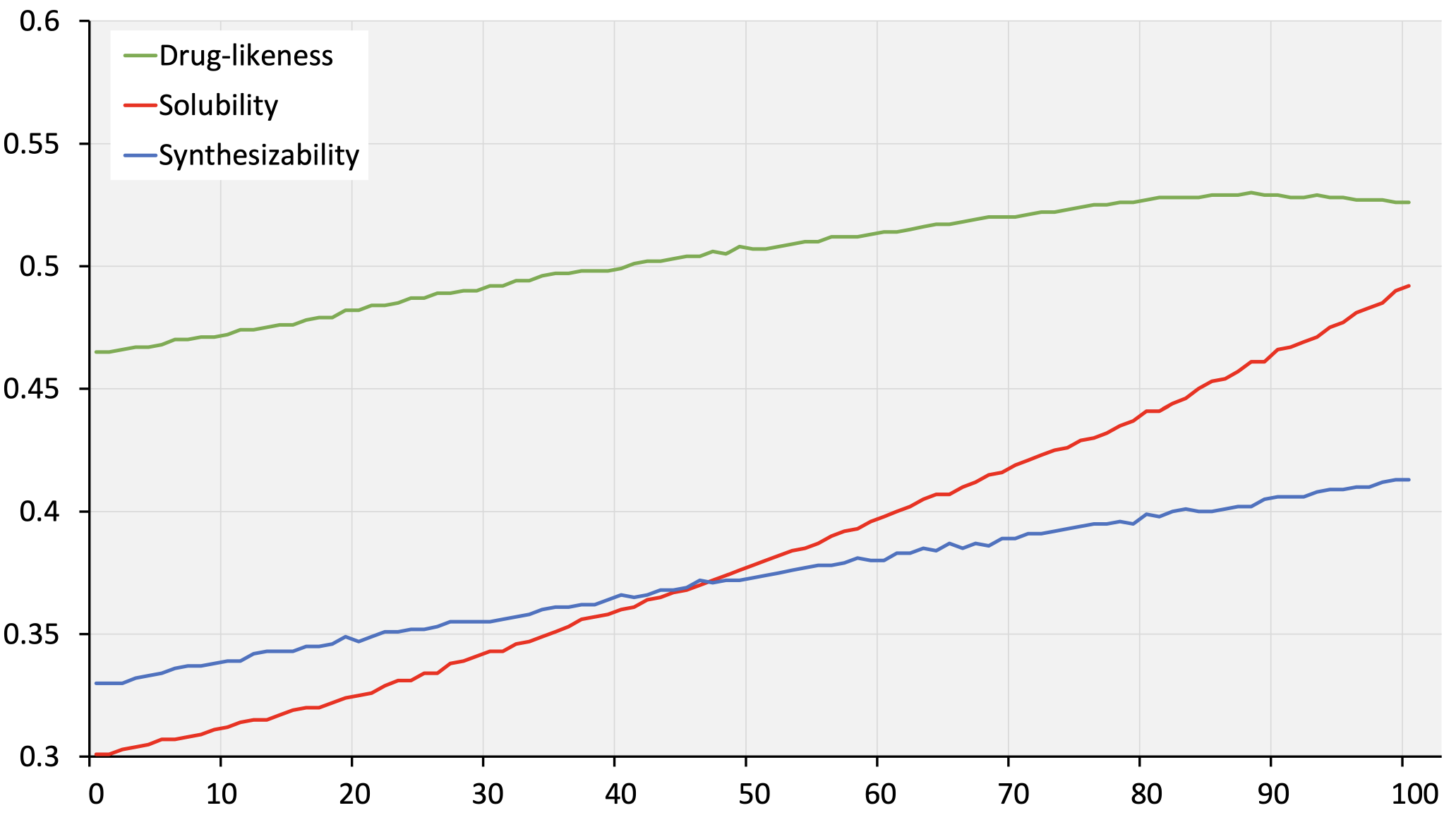}
\caption{Scaffold-based molecular generation.}
\end{subfigure}
\caption{Changes in the three property scores when training on the QM9 dataset.}
\label{fig:qm9_change}
\end{figure}   

\begin{figure}[ht]
\centering
\begin{subfigure}{0.42\textwidth}
\centering
\includegraphics[width=\textwidth]{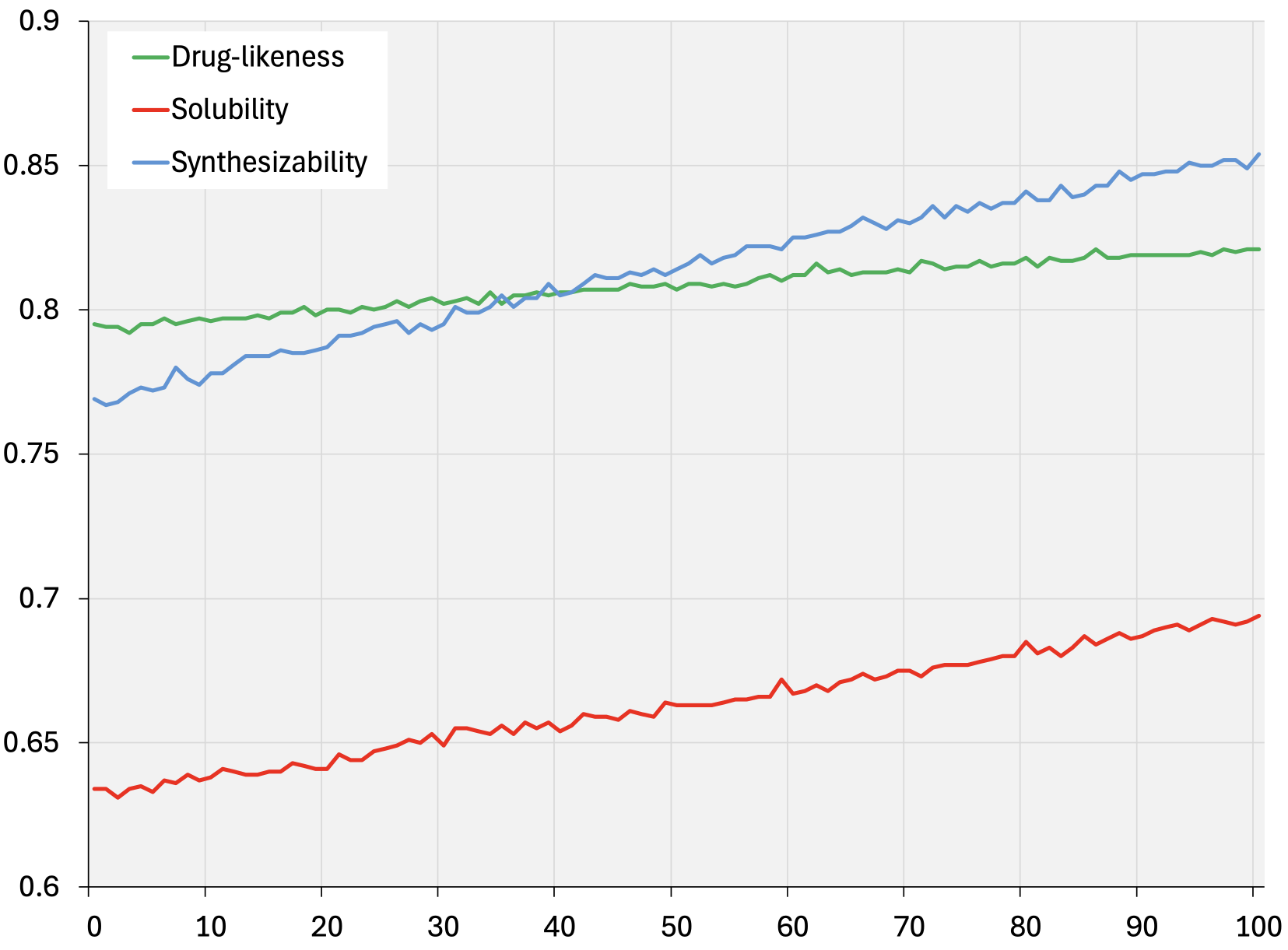}
\caption{$De~novo$ molecular generation.}
\end{subfigure}
\hspace{-4pt}
\begin{subfigure}{0.5\textwidth}
\centering
\includegraphics[width=\textwidth,height=0.62\hsize]{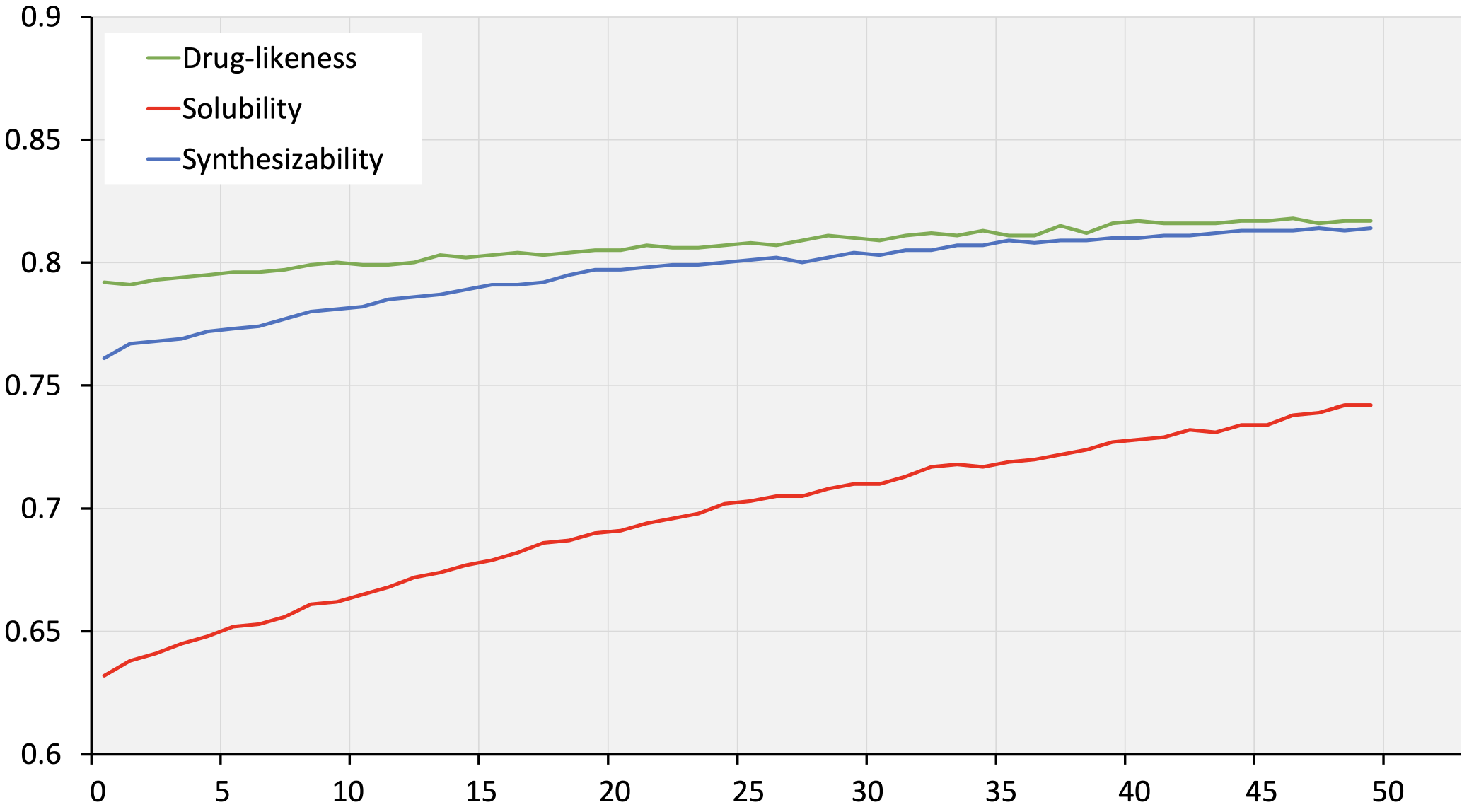}
\caption{Scaffold-based molecular generation.}
\end{subfigure}
\caption{Changes in the three property scores when training on the ZINC dataset.}
\label{fig:zinc_change}
\end{figure}   

\begin{figure}[htbp]
\begin{subfigure}{1\textwidth}
\centering
\includegraphics[width=0.8\textwidth]{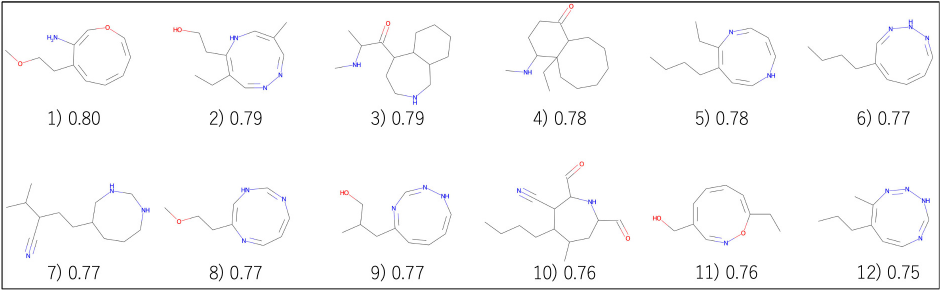}
\caption{Top-12 generated molecular structures with their QED scores on the QM9 dataset.}
\vspace{12pt}
\end{subfigure}

\begin{subfigure}{1\textwidth}
\centering
\includegraphics[width=0.8\textwidth]{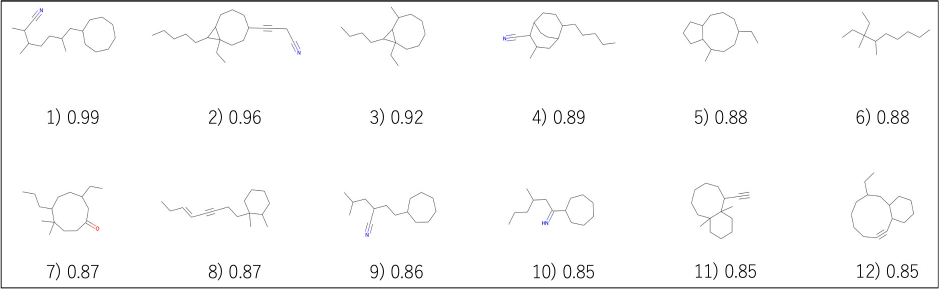}
\caption{Top-12 generated molecular structures with their logP scores on the QM9 dataset.}
\vspace{12pt}
\end{subfigure}

\begin{subfigure}{1\textwidth}
\centering
\includegraphics[width=0.8\textwidth]{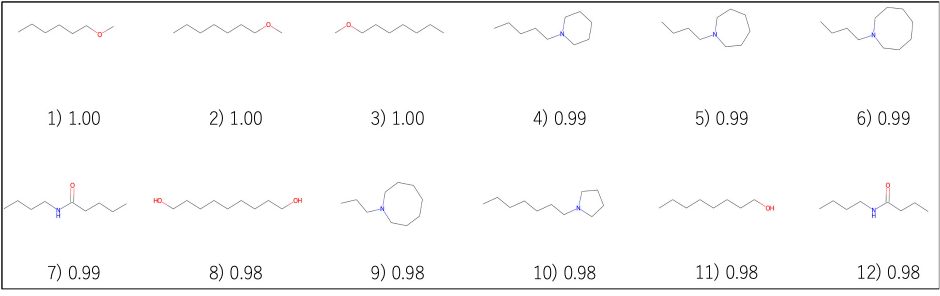}
\caption{Top-12 generated molecular structures with their SA scores on the QM9 dataset.}
\end{subfigure}
\caption{Top-12 molecular structures for $de~novo$ molecular generation on the QM9 dataset.}
\label{fig:top12_denovo_qm9}
\end{figure}

\begin{figure}[t]
\begin{subfigure}{1\textwidth}
\centering
\includegraphics[width=0.8\textwidth]{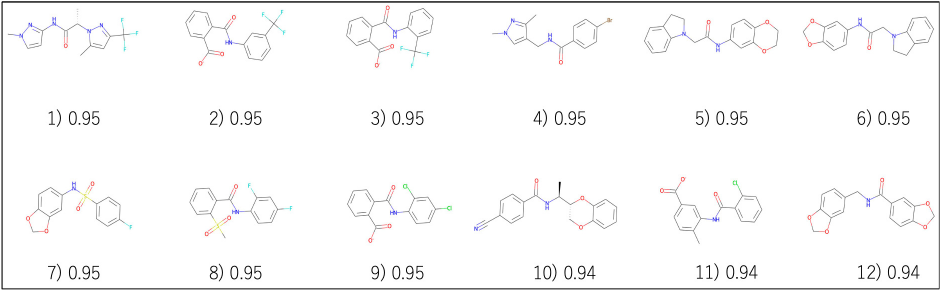}
\caption{Top-12 generated molecular structures with their QED scores on the ZINC dataset.}
\vspace{12pt}
\end{subfigure}

\begin{subfigure}{1\textwidth}
\centering
\includegraphics[width=0.8\textwidth]{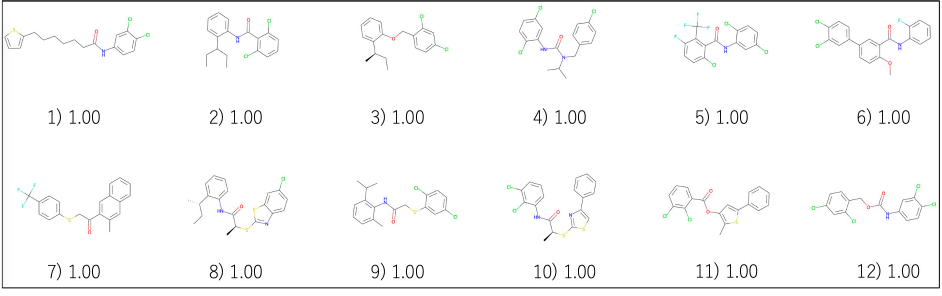}
\caption{Top-12 generated molecular structures with their logP scores on the ZINC dataset.}
\vspace{12pt}
\end{subfigure}

\begin{subfigure}{1\textwidth}
\centering
\includegraphics[width=0.8\textwidth]{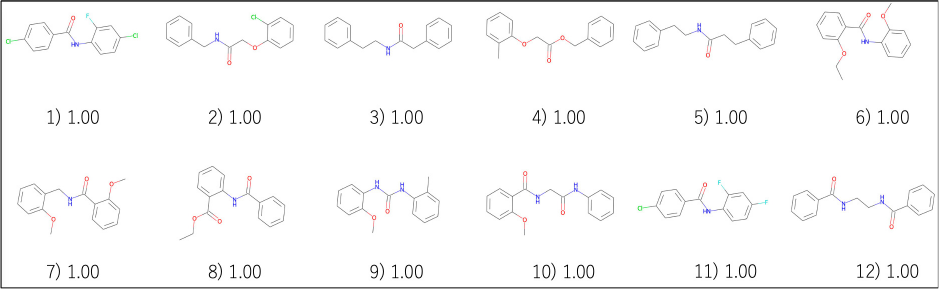}
\caption{Top-12 generated molecular structures with their SA scores on the ZINC dataset.}
\end{subfigure}

\caption{Top-12 molecular structures for $de~novo$ molecular generation on the ZINC dataset.}
\label{fig:top12_denovo_zinc}
\end{figure}

\begin{figure}[ht]
\begin{subfigure}{0.32\textwidth}
\centering
\includegraphics[width=1\textwidth]{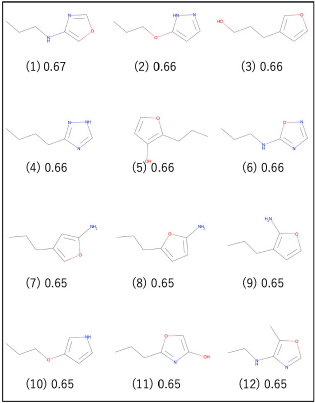}
\caption{Training dataset.}
\end{subfigure}
\hfill
\begin{subfigure}{0.32\textwidth}
\centering
\includegraphics[width=1\textwidth]{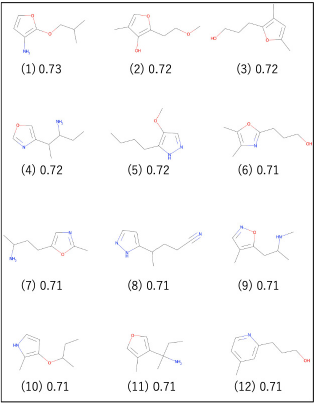}
\caption{RL-MolGAN.}
\end{subfigure}
\hfill
\begin{subfigure}{0.32\textwidth}
\centering
\includegraphics[width=1.04\hsize]{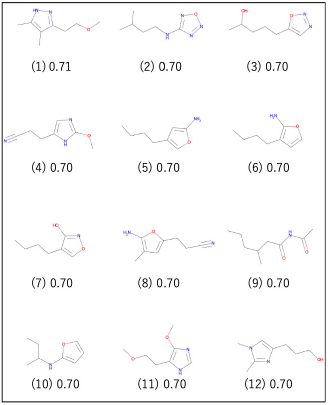}
\caption{RL-MolWGAN.}
\end{subfigure}
\caption{Top-12 molecular structures with their QED scores on the QM9 dataset for scaffold-based molecular generation.}
\label{fig:qm9_qed_12}
\end{figure}

\begin{figure}[htbp]
\begin{subfigure}{0.32\textwidth}
\centering
\includegraphics[width=1\textwidth]{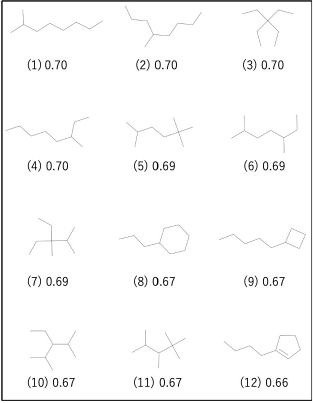}
\caption{Training dataset.}
\end{subfigure}
\hfill
\begin{subfigure}{0.32\textwidth}
\centering
\includegraphics[width=1\textwidth]{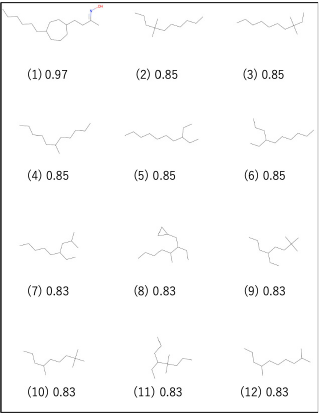}
\caption{RL-MolGAN.}
\end{subfigure}
\hfill
\begin{subfigure}{0.32\textwidth}
\centering
\includegraphics[width=1\textwidth]{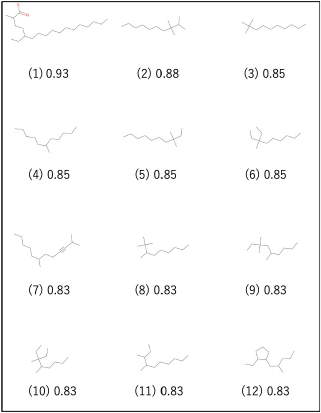}
\caption{RL-MolWGAN.}
\end{subfigure}
\caption{Top-12 molecular structures with their logP scores on the QM9 dataset for scaffold-based molecular generation.}
\label{fig:qm9_logp_12}
\end{figure}

\begin{figure}[htbp]
\begin{subfigure}{0.32\textwidth}
\centering
\includegraphics[width=1.02\textwidth]{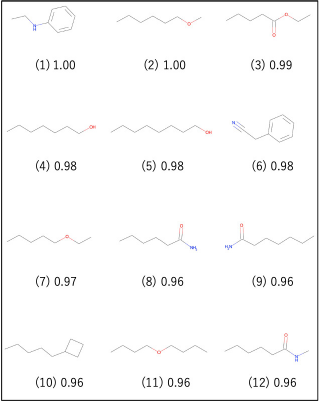}
\caption{Training dataset.}
\end{subfigure}
\hfill
\begin{subfigure}{0.32\textwidth}
\centering
\includegraphics[width=1\textwidth]{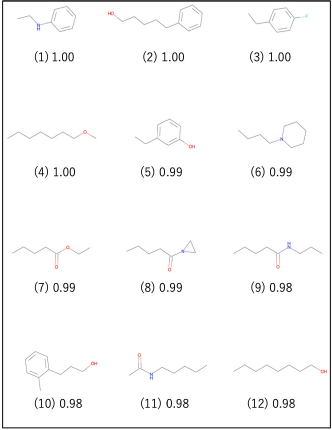}
\caption{RL-MolGAN.}
\end{subfigure}
\hfill
\begin{subfigure}{0.32\textwidth}
\centering
\includegraphics[width=1\textwidth]{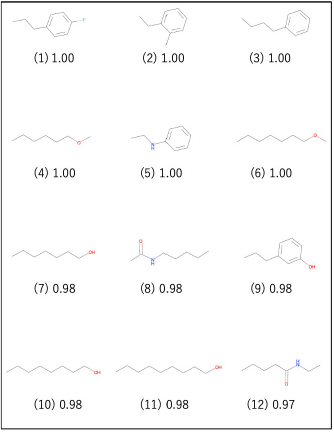}
\caption{RL-MolWGAN.}
\end{subfigure}
\caption{Top-12 molecular structures with their SA scores on the QM9 dataset for scaffold-based molecular generation.}
\label{fig:qm9_sa_12}
\end{figure}

\begin{figure}[htbp]
\begin{subfigure}{0.32\textwidth}
\centering
\includegraphics[width=1\textwidth]{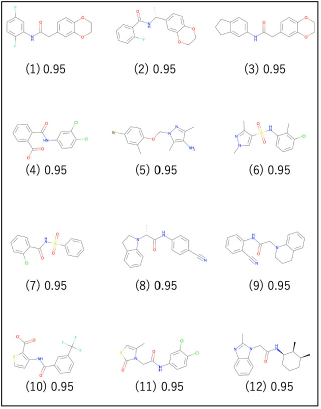}
\caption{Training dataset.}
\end{subfigure}
\hfill
\begin{subfigure}{0.32\textwidth}
\centering
\includegraphics[width=1\textwidth]{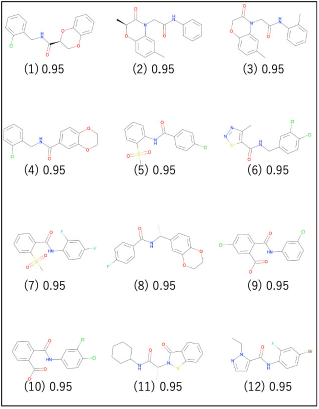}
\caption{RL-MolGAN.}
\end{subfigure}
\hfill
\begin{subfigure}{0.32\textwidth}
\centering
\includegraphics[width=1.02\textwidth]{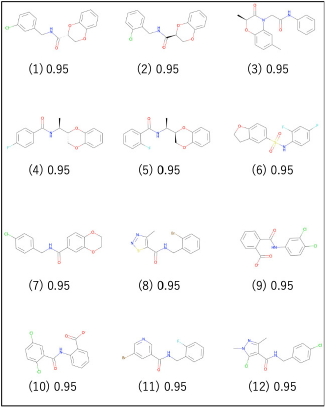}
\caption{RL-MolWGAN.}
\end{subfigure}
\caption{Top-12 molecular structures with their QED scores on the ZINC dataset for scaffold-based molecular generation.}
\label{fig:zinc_qed_12}
\end{figure}

\begin{figure}[htbp]
\begin{subfigure}{0.32\textwidth}
\centering
\includegraphics[width=1\textwidth]{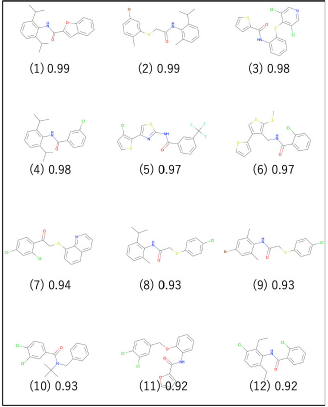}
\caption{Training dataset.}
\end{subfigure}
\hfill
\begin{subfigure}{0.32\textwidth}
\centering
\includegraphics[width=1\textwidth,height=1.25\hsize]{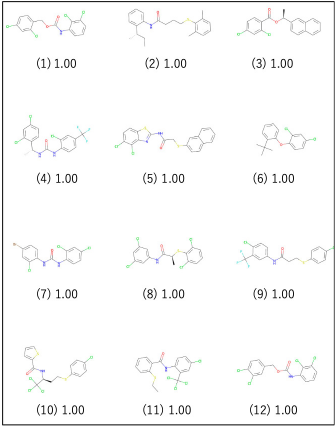}
\caption{RL-MolGAN.}
\end{subfigure}
\hfill
\begin{subfigure}{0.32\textwidth}
\centering
\includegraphics[width=1\textwidth,height=1.25\hsize]{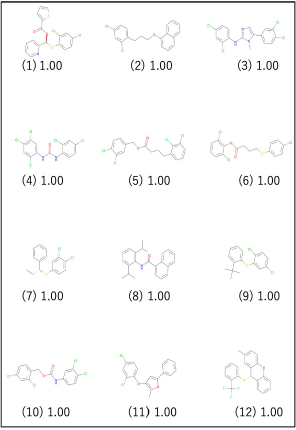}
\caption{RL-MolWGAN.}
\end{subfigure}
\caption{Top-12 molecular structures with their logP scores on the ZINC dataset for scaffold-based molecular generation.}
\label{fig:zinc_logp_12}
\end{figure}

\begin{figure}[htbp]
\begin{subfigure}{0.32\textwidth}
\centering
\includegraphics[width=1\textwidth]{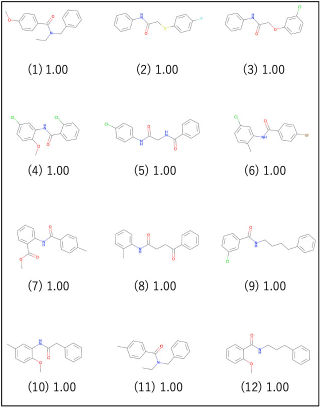}
\caption{Training dataset.}
\end{subfigure}
\hfill
\begin{subfigure}{0.32\textwidth}
\centering
\includegraphics[width=1\textwidth]{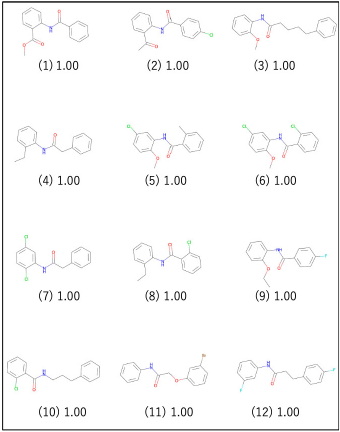}
\caption{RL-MolGAN.}
\end{subfigure}
\hfill
\begin{subfigure}{0.32\textwidth}
\centering
\includegraphics[width=1\textwidth,height=1.27\hsize]{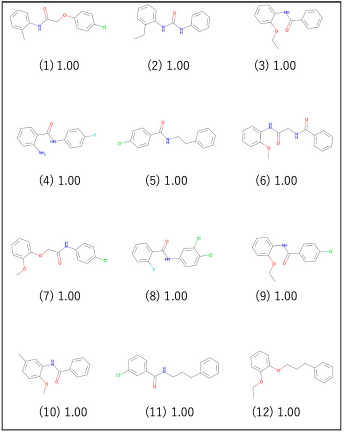}
\caption{RL-MolWGAN.}
\end{subfigure}
\caption{Top-12 molecular structures with their SA scores on the ZINC dataset for scaffold-based molecular generation.}
\label{fig:zinc_sa_12}
\end{figure}

\end{document}